\documentclass{article}
\usepackage{spconf,amsmath,graphicx}
\usepackage[linesnumbered,ruled,vlined]{algorithm2e}
\usepackage{url}
\RestyleAlgo{ruled}
\usepackage{amsmath}
\usepackage{amssymb}
\usepackage{multirow, multicol}
\usepackage{hyperref}
\usepackage[table,xcdraw,dvipsnames]{xcolor}
\usepackage{graphicx}
\usepackage{bm}
\usepackage{lipsum}
\usepackage{bbm}

\SetCommentSty{mycommfont}

\SetKwInput{KwInput}{Input}                % Set the Input
\SetKwInput{KwOutput}{Output}              % set the Output

\title{REVISITING HETEROPHILY IN GRAPH CONVOLUTIONAL NETWORKS BY LEARNING REPRESENTATIONS ACROSS TOPOLOGICAL AND FEATURE SPACES}

\name{Ashish Tiwari \qquad Sresth Tosniwal \qquad Shanmuganathan Raman}
  
\address{CVIG Lab, Indian Institute of Technology Gandhinagar \\ \{ashish.tiwari, sresth.t, shanmuga\}@iitgn.ac.in}
\begin{document}
\ninept
\maketitle
\begin{abstract}
Graph convolution networks (GCNs) have been enormously successful in learning representations over several graph-based machine learning tasks. Specific to learning rich node representations, most of the methods have solely relied on the homophily assumption and have shown limited performance on the heterophilous graphs. While several methods have been developed with new architectures to address heterophily, we argue that by learning graph representations across two spaces i.e., topology and feature space GCNs can address heterophily. In this work, we experimentally demonstrate the performance of the proposed GCN framework over semi-supervised node classification task on both homophilous and heterophilous graph benchmarks by learning and combining representations across the topological and the feature spaces. \footnotetext[1]{Code and additional implementation details can be found \href{https://sites.google.com/iitgn.ac.in/hetgcn/home}{here}.}
\end{abstract}
\section{Introduction}
\label{section:intro}
\footnotetext[2]{This work is supported by SERB MATRICS grant.} Owing to their great capacity in node representation learning, graph convolutional networks (GCNs) have widely been adopted across numerous graph-based machine learning tasks ranging from graph classification \cite{zhang2018end}, link prediction \cite{you2019position}, recommendation systems \cite{ying2018graph}, fault diagnosis \cite{chen2021graph}, network embedding \cite{fan2021propagation}, and computer vision \cite{zhao2019semantic}. However, most of the existing methods are centered around the homophily assumption for semi-supervised node classification (SSNC) task, i.e, a pair of nodes tend to be connected if they are similar or alike. The homophilous assumption is mainly attributed to the recursive neighbourhood aggregation mechanism in GCNs wherein a node's representation is obtained from its neighbours which are likely to have the same label. Several works \cite{zhu2020beyond,zhu2021graph} advocate that GNNs are best suited to address homophily and are not suitable for heterophilous graphs and have designed new architectures to explicitly address heterophily in graphs. However, Ma \emph{et al.} \cite{ma2021homophily} have shown that there exist certain specific types of heterophily over which GCNs in their original form can show strong performance. As per \cite{ma2021homophily}, GCNs learn equivalent embeddings for nodes that share same neighbourhood distribution. Further, Wang \emph{et al.} \cite{wang2020gcn} studied the performance of multi-channel graph convolutional networks in obtaining the information from both topological and features space. In this work, we take a step further and demonstrate that by learning and fusing features from topological and feature spaces, GCNs can address heterophily in a more general sense, i.e., not restricted to any specific kind of neighbourhood distributions.

\textbf{Homophily and heterophily in graphs.} When the connected nodes in a graph have similar labels or properties, the graph is said to have \textit{homophily}. Homophily is found in several real-world graphs such as friendship networks \cite{mcpherson2001birds}, political networks \cite{gerber2013political, newman2018networks}, citation networks \cite{ciotti2016homophily} and others. When the connected nodes have different properties or labels, the graph is said to possess \textit{heterophily}. The examples of heterophilous graphs include dating networks or molecular networks \cite{zhu2020beyond}. We observe that implicitly there is some sort of  commonality within a heterophilous connections. Consider a dating network, where potential dates are more likely to be connected if they share common interest or hobbies. Similarly, in molecular or protein networks, different types of amino acids make up a protein based on their bio-chemical compatibility. In both these examples, the common attributes between nodes of different labels shows inherent homophily (commonality) in a heterophilous connection.

\textbf{GCNs for heterophily.} Following these observations, we argue that GCNs can show improved performance over heterophilous graphs if they can be modeled to identify these hidden commonalities within a heterophilous connection. In traditional GCNs, the node representation learning is guided solely by topology graphs. However, both network topology and node features offer holistic understanding for the graph representation learning. We cannot obtain accurate neighborhood information by solely considering the topological view as the node features are often smoothened out when the feature information is propagated over network topological structure. Also, GCNs in their original form fail to adaptively learn correlation information between topological and feature views \cite{wang2020gcn}. In this work, we propose a multi-channel GCN framework that first extracts information individually from the topology and feature graph by using two separate GCN encoders and then uses another encoder to extract information common to both the spaces for the node classification task and apply this to address heterophily. Remember that the role of the common encoder is to fill in the missing information obtained from each of the individual spaces.
\vspace{-1mm}
\section{Related Work}
In this section, we briefly review the development of GCNs over the years and understand how they have been applied to address heterophily. Bruna \emph{et al.} \cite{bruna2013spectral} designed the graph convolution operation in Fourier domain using graph Laplacian and Defferrard \emph{et al.} \cite{defferrard2016convolutional} employed its Chebyshev expansion to improve the efficiency. Later, Kipf and Welling \cite{kipf2016semi} simplified the convolution operation and proposed feature aggregation from the one-hop neighbors. After this, a plenty of variants like GraphSAGE \cite{hamilton2017inductive}, GAT \cite{velivckovic2017graph}, SGC \cite{wu2019simplifying}, and GMNN \cite{qu2019gmnn}. were introduced. While they have have shown performance improvement over node classification, most of them have suffered due to inherent low-pass filtering of node features due to Laplacian smoothing. For a more detailed review on different variants of GCNs, we refer the readers to \cite{wu2020comprehensive, zhang2020deep}. Surprisingly, all these methods only use a single topology graph for node aggregation and fail to completely utilize the information from the feature space. A few methods like AM-GCN \cite{wang2020gcn} have shown better performance by extracting embeddings from both topology graph and feature graph. However, these have mostly performed well on homophilous graphs. Several recent works have methods to address heterophily in graphs by re-designing or modifying model architectures such as Geom-GCN \cite{pei2020geom}, H2GCN \cite{zhu2020beyond}, GPR-GNN \cite{chamberlain2021grand}, and CPGNN \cite{zhu2021graph}. Further, \cite{ma2021homophily} has established that GCN can potentially achieve good performance on some specific types of heterophily in graphs. In this work, our focus is on understanding whether the information learned and fused across topological and feature graphs can be used to address heterophily in general.

\section{Method}

\subsection{Mathematical Preliminaries.} Let us consider an undirected graph $\mathcal{G} = (\mathcal{V}, \mathcal{E})$ characterized by an adjacency matrix $\mathbf{A}$, a diagonal degree matrix $\mathbf{D}$ ($\mathbf{\widetilde{A}}$ and $\mathbf{\widetilde{D}}$ when self-loops are considered), $N$ nodes, and a node feature matrix $\mathbf{X} \in \rm I\!R^{N \times d}$ containing a $d$-dimensional feature vector for each node in the graph.

\textbf{GCN Layer.} A single GCN layer takes the form of $\mathbf{H}^{(l+1)} = \sigma(\widetilde{\mathbf{P}}\mathbf{H}^{(l)}\mathbf{W}^{(l)})$, where $\mathbf{H}^{(l)}$ and $\mathbf{H}^{(l+1)}$ represent the input and output features of the $l^{th}$ layer, $\mathbf{W}^{(l)}$ represents the parameter (weight) matrix that transforms features, and $\sigma$ represents ReLU activation. Further, $\widetilde{\mathbf{P}} = \widetilde{\mathbf{D}}^{-1/2}\widetilde{\mathbf{A}}\widetilde{\mathbf{D}}^{-1/2}$ is the degree normalized graph Laplacian matrix.

\textbf{Homophily.} The edge homophily ratio is defined as the fraction of edges that connect nodes with the same labels \cite{ma2021homophily}. Mathematically, 
\begin{equation}
    \centering
    \mathcal{H}(\mathcal{G},\mathcal{Y}) = \frac{1}{|\mathcal{E}|}\sum_{(i,j) \in \mathcal{E}}\mathbbm{1}(y_{i} = y_{j})
\end{equation}
Here, $\mathcal{Y}$ is the set of all the node labels $y_{i}$ and $\mathbbm{1}(\cdot)$ is an indicator function. $0.5 \leq \mathcal{H}(\cdot) \leq 1$ corresponds to high homophily and $0 \leq \mathcal{H}(\cdot) \leq 0.5$ corresponds to high heterophily.

\subsection{Proposed Framework} 
\begin{figure}[h]
    \centering
    \includegraphics[width=\linewidth]{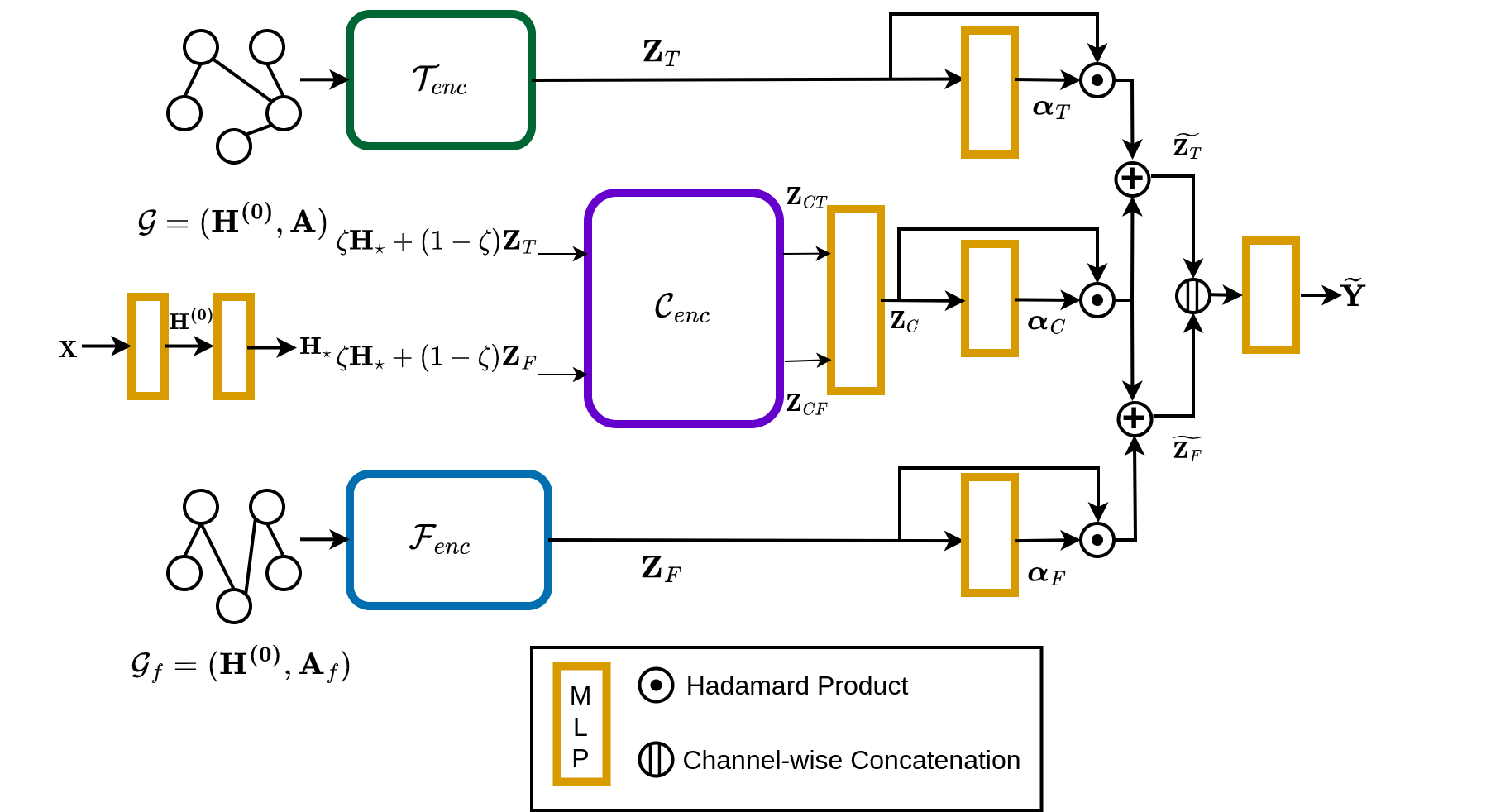}
    \caption{The architecture of the proposed framework.}
    \label{fig:1}
\end{figure}

The overall information flow in the proposed framework is shown in Figure \ref{fig:1}. The main idea is to propagate the node features through both the topological and the feature space. We start with two 2-layer GCN encoders $\mathcal{T}_{enc}$ and $\mathcal{F}_{enc}$ that take $\mathcal{G}=(\mathbf{H^{(0)}}, \mathbf{A})$ and $\mathcal{G}_{f}=(\mathbf{H^{(0)}}, \mathbf{A}_{f})$ as input, respectively (with $\mathbf{H^{(0)}}=f_{MLP}(\mathbf{X})$), to obtain the feature maps $\mathbf{Z}_{T}$ and $\mathbf{Z}_{F}$ corresponding to two views. Please note that $\mathcal{G}$ is the regular topological graph and $\mathcal{G}_{f}$ is the k-nearest neighbour feature graph based on node feature matrix $\mathbf{X}$ and adjacency matrix $\mathbf{A}_{f}$ of the k-NN graph. \footnotetext[3]{We use cosine similarity to compute the similarity $s_{ij}$ between two node features $\mathbf{x}_{i}$ and $\mathbf{x}_{j}$.
$s_{ij} = \frac{\mathbf{x}_{i}.\mathbf{x}_{j}}{|\mathbf{x}_{i}||\mathbf{x}_{j}|}$}

Further, each GCN layer is designed to have a residual connection to initial features $\mathbf{H}^{(0)}$ (or $\mathbf{X}$), as described in Equation \ref{eq:2}.
\begin{equation}
    \centering
    \mathbf{H}^{(l+1)} = \alpha \mathbf{H}^{(l)} + (1-\alpha) \mathbf{H}^{(0)} 
    \label{eq:2}
\end{equation}
Here, $\alpha = 0.8$. We deploy a residual connection from the input features $\mathbf{H}^{(0)}$ such that the final representation of each node retains at least a fraction of information $(\alpha)$ from the input features. This helps to slow down the rate of convergence of features to a point or smaller subspace - a common cause of oversmoothing in GCNs \cite{chen2020simple,tiwari2022exploring}.

As discussed in Section \ref{section:intro}, we learn the information with common characteristics between these two spaces. We deploy a common GCN encoder $\mathcal{C}_{enc}$ with shared parameter to learn the embeddings $\mathbf{Z}_{CT}$ and $\mathbf{Z}_{CF}$ which in turn are combined together to get the common embedding $\mathbf{Z}_{C}$ using a multi-layer perceptron (MLP). However, unlike methods like AM-GCN \cite{wang2020gcn}, we use a weighted combination of the learned features ($\mathbf{Z}_{T}$ and $\mathbf{Z}_{F}$) and the initial features $\mathbf{X}$ as input to $\mathcal{C}_{enc}$. Specifically, $    \mathbf{Z}_{{C}_{in}} = \zeta \mathbf{H_{\star}} + (1-\zeta) \mathbf{Z}_{\star}$ Here, $\mathbf{H_{\star}} = f_{MLP}(\mathbf{H^{0}})$ and $\mathbf{Z}_{\star}$ represents either $\mathbf{Z}_{T}$ or $\mathbf{Z}_{F}$ and we set $\zeta=0.85$ This formulation was found to provide performance improvement over just using the learned $\mathbf{Z}_{T}$ or $\mathbf{Z}_{F}$.

With $\mathbf{Z}_{T}$, $\mathbf{Z}_{F}$, and $\mathbf{Z}_{C}$ at hand, we now wish to find an optimal combination of these features that decides the final predicted labels, as governed by Equation \ref{eq:3}.
\begin{equation*}
    \centering
    \widetilde{\mathbf{Z}_{T}} = \boldsymbol{\alpha}_{T} \odot \mathbf{Z}_{T} + \boldsymbol{\alpha}_{C} \odot \mathbf{Z}_{C}
\end{equation*}
\begin{equation}
    \centering
    \widetilde{\mathbf{Z}_{F}} = \boldsymbol{\alpha}_{F} \odot \mathbf{Z}_{F} + \boldsymbol{\alpha}_{C} \odot \mathbf{Z}_{C}
    \label{eq:3}
\end{equation}
Here, $\boldsymbol{\alpha}_{T}$, $\boldsymbol{\alpha}_{F}$, and $\boldsymbol{\alpha}_{C}$ are the feature-level attention weights for $\mathbf{Z}_{T}$, $\mathbf{Z}_{F}$, and $\mathbf{Z}_{C}$, respectively, such that $\boldsymbol{\alpha}_{\star} = f_{\star MLP}(Z_{\star})$ where $\star\in \{T, F, C\}$, $f_{\star MLP}$ is the corresponding MLP, and $\odot$ is the element-wise Hadamard product. Consequently, we obtain the final set of labels $\widehat{\mathbf{Y}}$ as shown in Equation \ref{eq:4}.
\begin{equation}
    \centering
    \widehat{\mathbf{Y}} = softmax(\mathbf{W}.\widehat{\mathbf{Z}} + \mathbf{b})
    \label{eq:4}
\end{equation}
where, $\widehat{\mathbf{Z}} = (\widetilde{\mathbf{Z}_{T}} || \widetilde{\mathbf{Z}_{F}})$.

\vspace{-1mm}
\subsection{Training Details}
We train and evaluate the proposed framework over both homophilous and heterophilous graph datasets as listed in Table \ref{table:1}. 

\textbf{Closeness Constraint.} We want the two output embeddings $\mathbf{Z}_{CT}$ and $\mathbf{Z}_{CF}$ arising out of $\mathcal{C}_{enc}$ to be close as possible. Therefore, we apply a closeness constraint enabling to search for commonality. Following \cite{wang2020gcn}, we first obtain the similarity of $N$ nodes as,
\begin{equation*}
    \centering
    \mathbf{S}_{T} = \mathbf{Z}_{CTnor}.{\mathbf{Z}^{T}}_{CTnor}
\end{equation*}
\begin{equation}
    \centering
    \mathbf{S}_{F} = \mathbf{Z}_{CFnor}.{\mathbf{Z}^{T}}_{CFnor}
\end{equation}
Here, $\mathbf{Z}_{CTnor}$ and $\mathbf{Z}_{CFnor}$ are the $L_{2}$-normalized version of $\mathbf{Z}_{CT}$ and $\mathbf{Z}_{CF}$, respectively. Using these, we define the closeness constraint using the Frobenious norm, as shown in Equation \ref{eq:6}.

\begin{equation}
    \centering
    \mathcal{L}_{c} = ||\mathbf{S}_{T} - \mathbf{S}_{F}||^{2}_{F}
    \label{eq:6}
\end{equation}

\textbf{Disparity Constraint.} An input graph from the same view generates two representations, i.e., $\mathcal{G} \rightarrow  (\mathbf{Z}_{T}, \mathbf{Z}_{CT})$ and $\mathcal{G}_{f} \rightarrow (\mathbf{Z}_{F}, \mathbf{Z}_{CF})$. In order to ensure that they capture different information, we introduce a disparity regularization $\mathcal{L}_{d}$ (similar to \cite{fan2021propagation}) to enlarge distances between them using the negative cosine similarity, as described in Equation \ref{eq:7}.
\begin{equation}
    \centering
    \mathcal{L}_{d} = -\frac{1}{N}\sum_{i=1}^{N}\Biggl( \frac{\mathbf{z}_{t}^{(i)}.\mathbf{z}_{ct}^{(i)}}{||\mathbf{z}_{t}^{(i)}||.||\mathbf{z}_{ct}^{(i)}||} + \frac{\mathbf{z}_{f}^{(i)}.\mathbf{z}_{cf}^{(i)}}{||\mathbf{z}_{f}^{(i)}||.||\mathbf{z}_{cf}^{(i)}||}\Biggl)
    \label{eq:7}
\end{equation}

\textbf{Classification Loss.} Let $\mathbf{Y}$ be the one-hot encoded label matrix, $\mathcal{V}_{train}$ be the set of training nodes, and $\widehat{\mathbf{Y}}$ be the prediction, as defined in Equation \ref{eq:4}. We minimize the standard cross-entropy loss to train the network.
\begin{equation}
    \centering
    \mathcal{L}_{cl} = -\sum_{v \in \mathcal{V}_{train}}\mathbf{Y}ln\widehat{\mathbf{Y}}
    \label{eq:8}
\end{equation}
The overall optimization objective function is as described in Equation \ref{eq:9}.
\begin{equation}
    \mathcal{L}_{T} = \lambda\mathcal{L}_{cl} + \beta\mathcal{L}_{c} + \gamma\mathcal{L}_{d} 
    \label{eq:9}
\end{equation}
Here, $\lambda, \beta$, and $\gamma$ are the weights for the classification loss, closeness, and disparity constraints, respectively. \footnote{Due to space constraints, we provide additional implementation details including the hyper-parameter values in the associated config files \href{https://sites.google.com/iitgn.ac.in/hetgcn/home}{here}.}

\section{Experimental Evaluation}
In this section, we evaluate the proposed approach on homophilous and heterophilous graph datasets (see Table \ref{table:1}) and reason out several design choices. Please note that we have chosen high performing standard GCN variants and AM-GCN \cite{wang2020gcn} for a fair comparison keeping in mind that our aim is to establish the ability of GCNs in addressing heterophily with any major change in its original formulation. All experiments have been performed with $40$ labeled nodes per class. We find that this is a reasonable approximation of the real world scenario where we are less likely to provided with lots of labeled nodes.

\begin{table}[h]
\centering
\resizebox{\linewidth}{!}{%
\begin{tabular}{c|cccc|c}
\hline
Dataset & \# Classes & \# Nodes & \# Edges & \# Features & Heterophily \\ \hline
ACM            & 3  & 3,025 & 13,128  & 1,870 & 0.179 \\
Citeseer         & 6  & 3,327 & 4,732  & 3,703  & 0.253 \\
CoraFull         & 70 & 19,793 & 65,311 & 8710 & 0.412\\ \hline
BlogCatalog & 6  & 5,196  & 1,71,743 & 8,189 & 0.599 \\ 
UAI2010     & 19  & 3,067  & 28,311 & 4,973 & 0.636\\
Flickr  & 9  & 7,575 & 2,39,738 & 12,047 & 0.761 \\\hline
\end{tabular}}
\caption{Dataset Statistics.}
\label{table:1}
\end{table}

\subsection{Semi-supervised node classification}
Table \ref{table:2} shows the performance of the proposed approach with the most recent and relevant baselines over six different graph datasets in terms of accuracy and the F1-score. We observe that the proposed framework either performs best or the second best among different baselines. It achieves relative improvements of  around $3.78\%$ (accuracy) and $4.61\%$ (F1-score) averaged over the heterophilous datasets. Further, the proposed framework is observed to outperform GCN and kNN-GCN on nearly all the datasets. This indicates the effectiveness of extracting useful information from both the spaces rather than only from the topological space (as in GCN) or the feature space (as in kNN-GCN). Moreover, we find that GCN operating solely on a topology graph fails to provide better result than on a feature space graph. This is due to the inherent useful differences in information that each space carries, thus establishing the need of introducing feature space graph in GCN. Overall, the margin of improvement over heterophilous datasets is observed to be higher than that over the homophilous datasets. This establishes our claim that learning representation across topological and feature spaces does help in better addressing the heterophily. 

\textbf{Network design choices}
We discuss the three important design choices in the proposed framework. (a) \textit{Initial Residual Connection}: We observed an improvement of $2.14\%$ (accuracy) and $3.09\%$ (F1-score) on an average by using initial residual connection (Equation \ref{eq:2}) in GCNs versus vanilla GCN. (b).\textit{ Feature-wise attention values}: Owing to a better feature selectivity for node classification \cite{tiwari2022exploring}, we observed improved performance by learning feature-wise attention values (weighing each feature of every node) instead of just one attention value per node (as AM-GCN\cite{wang2020gcn}). Finally, (c) \textit{Negative cosine similarity as disparity constraint}: It speeds up (almost $\times4$) the training (and offers better performance) when compared to Hilbert-Schmidt Independence Crtierion (HSIC), as used in \cite{wang2020gcn}. 

\begin{table}[h]
\centering
\resizebox{\linewidth}{!}{%
\begin{tabular}{c|ccccccc}
\hline
Datasets                     & Metrics & GCN         & kNN-GCN & GAT            & MixHop & AM-GCN         & Ours           \\\hline
\multirow{2}{*}{ACM}         & Acc     & 89.06       & 81.66   & 88.60          & 82.34  & \textbf{90.76} & \underline{90.71}    \\
                             & F1      & 89.00       & 81.53   & 88.55          & 81.13  & \underline { 90.66}    & \textbf{90.75} \\ \hline
\multirow{2}{*}{Citeseer}    & Acc     & \underline{73.10} & 61.54   & 73.04          & 71.48  & \textbf{74.70} & \textbf{74.70} \\
                             & F1      & 69.70       & 59.33   & 69.58          & 67.40  & \underline{69.81}    & \textbf{70.39} \\\hline
\multirow{2}{*}{CoraFull}    & Acc     & 58.12       & 44.80   & \textbf{60.61} & 57.20  & 58.60          & \underline{59.90}    \\
                             & F1      & 54.87       & 40.42   & \textbf{56.43} & 53.55  & 54.54          & \underline{55.31}    \\\hline
\multirow{2}{*}{BlogCatalog} & Acc     & 71.28       & 80.84   & 67.40          & 71.66  & \underline{84.94}    & \textbf{88.83} \\
                             & F1      & 70.71       & 80.16   & 66.39          & 70.84  & \underline{84.32}    & \textbf{88.39} \\\hline
\multirow{2}{*}{UAI2020}     & Acc     & 51.80       & 68.74   & 63.74          & 65.05  & \underline{73.40}    & \textbf{76.12} \\
                             & F1      & 33.80       & 54.45   & 45.08          & 53.86  & \underline { 62.19}    & \textbf{64.46} \\\hline
\multirow{2}{*}{Flickr}      & Acc     & 45.48       & 75.08   & 38.44          & 55.19  & \underline{79.81}    & \textbf{82.90} \\
                             & F1      & 43.27       & 75.40   & 36.94          & 56.25  & \underline{78.60}    & \textbf{82.82}\\\hline
                             
\end{tabular}%
}
\caption{Node classification results in terms of Accuracy and F1-score across different real-world datasets. Best performance is reported in \textbf{BOLD} and second best performance is ("\underline{underlined}").}
\label{table:2}
\end{table}

\begin{algorithm}[h]
\DontPrintSemicolon
  
  \KwInput{$\mathcal{G}=\{\mathcal{V},\mathcal{E}\}$, $\widehat{\mathcal{H}}$, $N$ and $\mathcal{Y}=\{y_{1}, y_{2},...,y_{\mathcal{C}}\}$}
  \KwOutput{$\mathcal{G'}=\{\mathcal{V},\mathcal{E'}\}$}
  Initialise $\mathcal{G'}=\{\mathcal{V},\mathcal{E}\}$ and $\widehat{\mathcal{H}}= h_{init}$; the original heterophily of the dataset.\;
  Obtain $L$; the list containing $10$ heterophily ratios such that $h_{init} \leq \widehat{\mathcal{H}} \leq 0.95$ \;
  
    \For{$\widehat{\mathcal{H}} \in L$}
    {
        Compute $K$, the number of heterophilous edges to added\;
   		Sample $K$ nodes such that each node $i \sim Uniform(\mathcal{V})$ \;
   		
   		\For {$i=1$ to $K$}
   		    {
   		    Obtain the label $y_{i}$ for node $i$ \;
   		    Sample a label $y_{j}\in \mathcal{Y}-y_{i}$ with probability $N_{j}/N$ \;
   		    Sample a node $j \sim Uniform(\mathcal{V}_{j})$ \;
   		    Update the edge set $\mathcal{E'} = \mathcal{E'} \cup \{(i,j)\}$
   		    
   		    }
   		    
    }
    \Return $\mathcal{G'}=\{\mathcal{V},\mathcal{E'}\}$

\caption{Adding Heterophilous Edges}
\label{alg:1}
\end{algorithm}

\subsection{Effect of extent of heterophily}
In this section, we attempt to understand the effect of variation in the extent of heterophily $\widehat{\mathcal{H}}$ over the model performance. We synthetically alter the topological graph $\mathcal{G}=(\mathbf{X}, \mathbf{A})$ starting from the original heterophily $h_{init}$ up-to $95\%$ heterophily in all the six graph datasets as per Algorithm \ref{alg:1}. Here, $N=|\mathcal{V}|$ and $\mathcal{V}_{j}$ is the set of nodes with label $y_{j}$.

\begin{figure}[h]
    \centering
    \includegraphics[width=\linewidth]{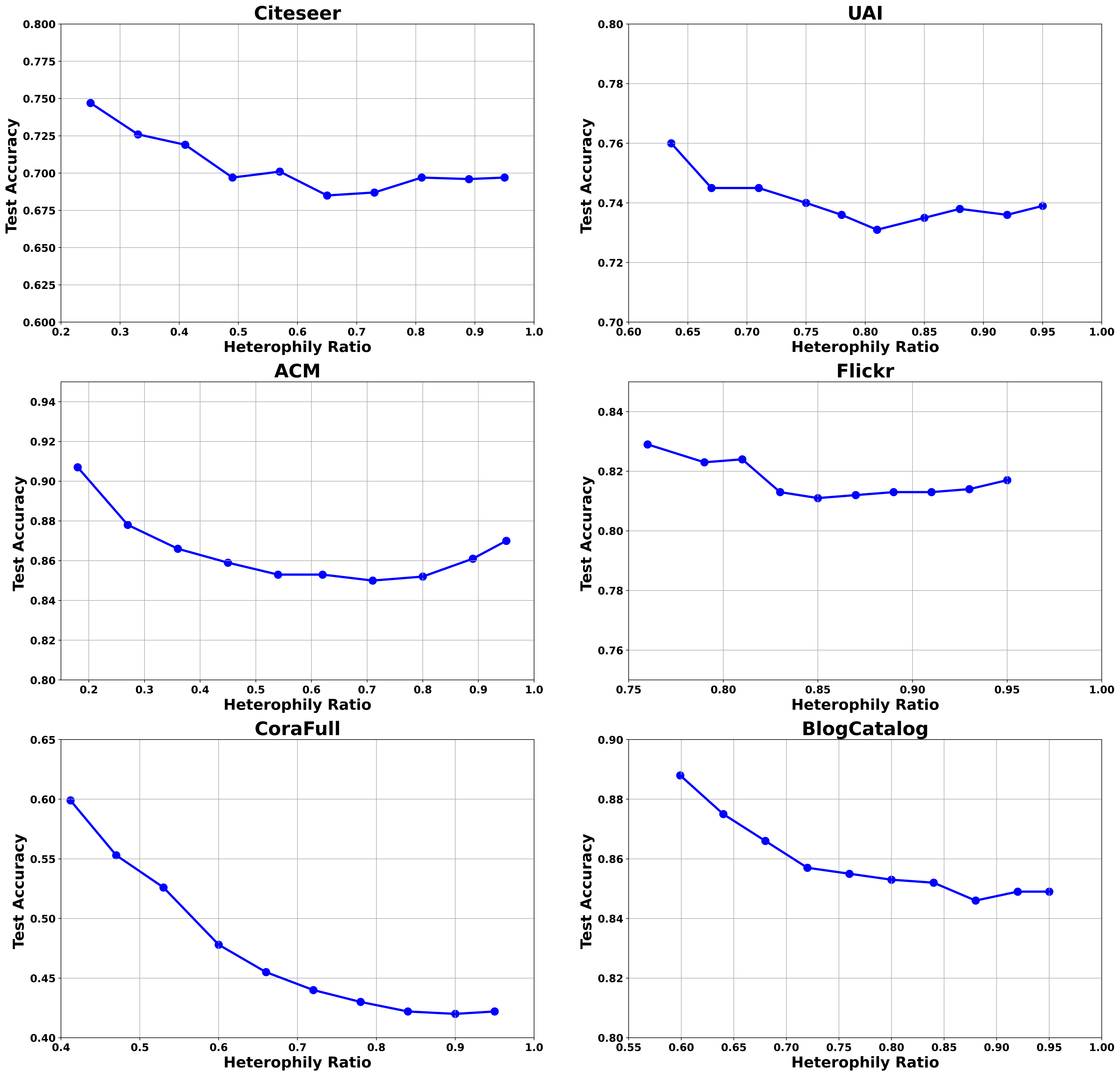}
    \caption{Effect of increasing heterophily on homophilous (first column) and heterophilous (second column) datasets.}
    \label{fig:2}
\end{figure}

Figure \ref{fig:2} shows the performance variation with increasing heterophily. Each point on plot in Figure \ref{fig:2} represents the performance of the proposed framework on the generated graph corresponding to the heterophily value in x-axis with the first point denoting the original graph, i.e, $K = 0$. As depicted in Figure \ref{fig:2}, as $K$ increases (thus, the heterophily) the classification performance first decreases and then eventually begins to increase (except on the CORA dataset). This shows that the the proposed GCN framework can perform under higher heterophily as well. The shape of the curve is attributed to the “phase transition” where at the beginning the initial topology is disturbed by added edges making the topological space less informative. However, as we add more edges, the network adapts itself to look for heterophilous properties in the feature space and starts showing better performance. While this is similar to that observed in \cite{ma2021homophily} for specific kind of heterophily, our framework applies to a general case by adapting itself to increasing heterophily. However, the performance on CORA dataset reduces as $K$ increases. This can be attributed to the fact that randomly adding heterophilous edges to a highly homophilous graph can sometimes not be very meaningful and can turn out to be noisy, similar to adding an edge between nodes that are actually never related in real-world.

\subsection{Analysis of attention values}

\begin{figure}[h]
    \centering
    \includegraphics[width=\linewidth]{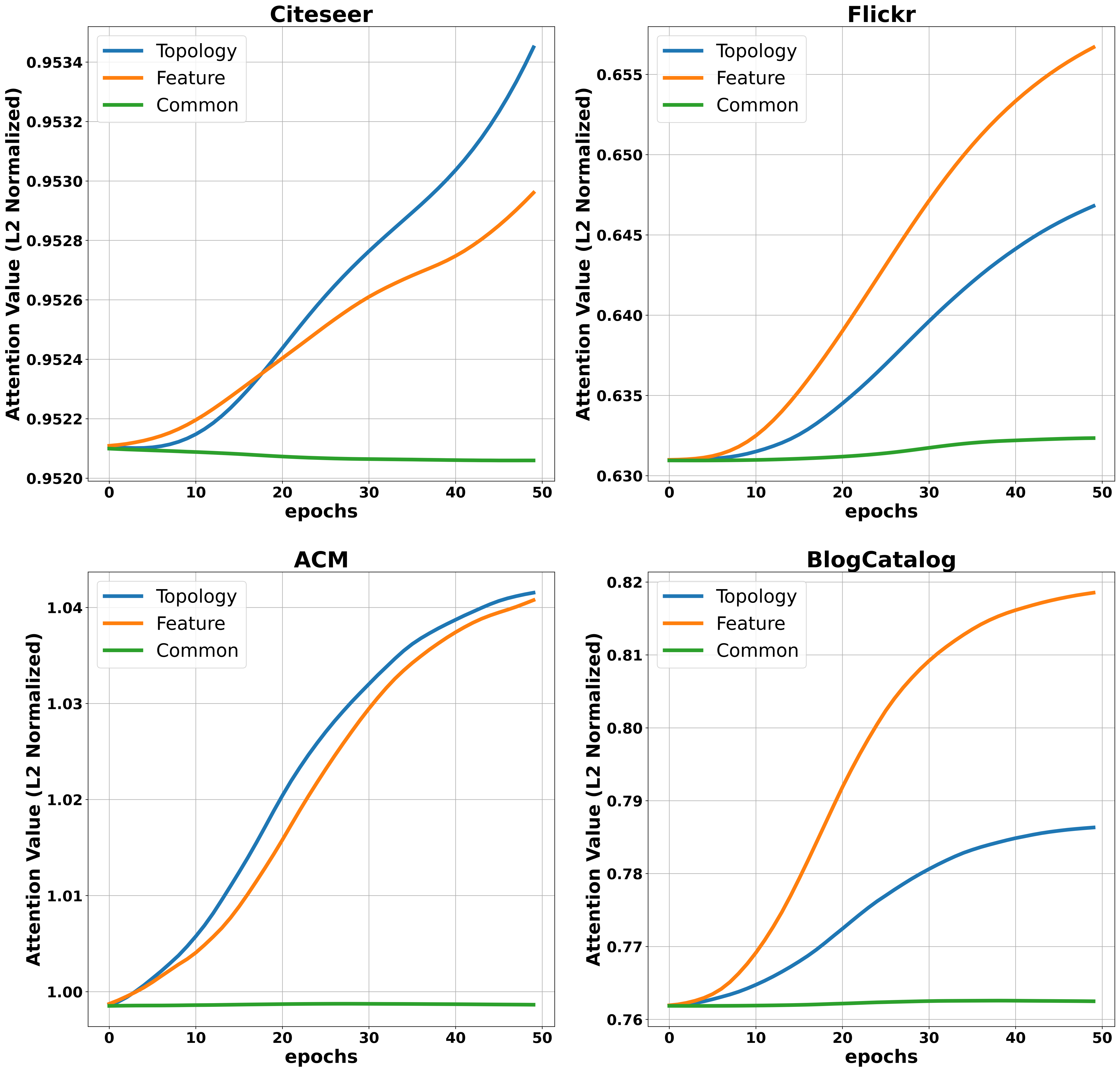}
    \caption{Variation of L2-normalised attention values learned over topological and feature spaces.}
    \label{fig:3}
\end{figure}

In this section, we analyze the variation in attention values during the training process. The main idea is to get an understanding of how does the framework decide to choose between the information from topological and feature space under different extent of heterophily. Here, we consider ACM and Citeseer (homophilous) and Flickr and BlogCatalog (heterophilous) datasets for visualization. Since the attention maps $(\boldsymbol{\alpha}_{T})$, $(\boldsymbol{\alpha}_{F})$, and $(\boldsymbol{\alpha}_{C})$ are matrices, we first L2-normalize them across rows and then compute the average of these row-wise norms. Figure \ref{fig:3} shows that at the beginning, the attention values (in the L2-norm sense) of topology $(\boldsymbol{\alpha}_{T})$, feature $(\boldsymbol{\alpha}_{F})$, and common $(\boldsymbol{\alpha}_{C})$ embeddings are almost the same. However, as the number of epochs increase, the attention values begin to differ. Specifically, for homophilous datasets (Figure \ref{fig:3}, first column), $(\boldsymbol{\alpha}_{T})$ increases more than $(\boldsymbol{\alpha}_{F})$ and vice-versa for the heterophilous datasets (Figure \ref{fig:3}, second column). Following these observations, we see that for homophilous datasets, the framework focuses more on information from the topological space and that on the feature space over heterophilous datasets. The information from both the spaces augment network learning by adaptively assigning appropriate attention values to information from each space and perform well on both homophilous and heterophilous datasets.

\vspace{-1mm}

\section{Conclusion}
We explore the ability of GCN in fusing information from network topological and feature space, and additionally address heterophily in graphs. While there is a method that shows GCNs in their original form addressing certain specific type of heterophily, others have advocated that GCNs are implicitly designed with homophily in mind. Further, they have not been suitable for heterophilous graphs unless their architectures are modified and redesigned specifically to address heterophily. We have shown that the proposed framework is able to handle heterophily across different real-world datasets with varied amount of heterophily not restricted solely to certain specific types. In the near future, we would like to explore whether such a multi-channel framework can be used to predict the missing node features in a graph and extend to dynamic graphs.

\newpage
% -------------------------------------------------------------------------
\bibliographystyle{IEEEbib}
{\footnotesize \bibliography{strings}}

\begin{thebibliography}{10}

\bibitem{zhang2018end}
Muhan Zhang, Zhicheng Cui, Marion Neumann, and Yixin Chen,
\newblock ``An end-to-end deep learning architecture for graph
  classification,''
\newblock in {\em Proceedings of the AAAI conference on artificial
  intelligence}, 2018, vol.~32.

\bibitem{you2019position}
Jiaxuan You, Rex Ying, and Jure Leskovec,
\newblock ``Position-aware graph neural networks,''
\newblock in {\em International conference on machine learning}. PMLR, 2019,
  pp. 7134--7143.

\bibitem{ying2018graph}
Rex Ying, Ruining He, Kaifeng Chen, Pong Eksombatchai, William~L Hamilton, and
  Jure Leskovec,
\newblock ``Graph convolutional neural networks for web-scale recommender
  systems,''
\newblock in {\em Proceedings of the 24th ACM SIGKDD International Conference
  on Knowledge Discovery \& Data Mining}, 2018, pp. 974--983.

\bibitem{chen2021graph}
Zhiwen Chen, Jiamin Xu, Tao Peng, and Chunhua Yang,
\newblock ``Graph convolutional network-based method for fault diagnosis using
  a hybrid of measurement and prior knowledge,''
\newblock {\em IEEE transactions on cybernetics}, 2021.

\bibitem{fan2021propagation}
Xiaolong Fan, Maoguo Gong, Yue Wu, AK~Qin, and Yu~Xie,
\newblock ``Propagation enhanced neural message passing for graph
  representation learning,''
\newblock {\em IEEE Transactions on Knowledge and Data Engineering}, 2021.

\bibitem{zhao2019semantic}
Long Zhao, Xi~Peng, Yu~Tian, Mubbasir Kapadia, and Dimitris~N Metaxas,
\newblock ``Semantic graph convolutional networks for 3d human pose
  regression,''
\newblock in {\em Proceedings of the IEEE/CVF Conference on Computer Vision and
  Pattern Recognition}, 2019, pp. 3425--3435.

\bibitem{zhu2020beyond}
Jiong Zhu, Yujun Yan, Lingxiao Zhao, Mark Heimann, Leman Akoglu, and Danai
  Koutra,
\newblock ``Beyond homophily in graph neural networks: Current limitations and
  effective designs,''
\newblock {\em Advances in Neural Information Processing Systems}, vol. 33, pp.
  7793--7804, 2020.

\bibitem{zhu2021graph}
Jiong Zhu, Ryan~A Rossi, Anup Rao, Tung Mai, Nedim Lipka, Nesreen~K Ahmed, and
  Danai Koutra,
\newblock ``Graph neural networks with heterophily,''
\newblock in {\em Proceedings of the AAAI Conference on Artificial
  Intelligence}, 2021, vol.~35, pp. 11168--11176.

\bibitem{ma2021homophily}
Yao Ma, Xiaorui Liu, Neil Shah, and Jiliang Tang,
\newblock ``Is homophily a necessity for graph neural networks?,''
\newblock {\em arXiv preprint arXiv:2106.06134}, 2021.

\bibitem{wang2020gcn}
Xiao Wang, Meiqi Zhu, Deyu Bo, Peng Cui, Chuan Shi, and Jian Pei,
\newblock ``Am-gcn: Adaptive multi-channel graph convolutional networks,''
\newblock in {\em Proceedings of the 26th ACM SIGKDD International conference
  on knowledge discovery \& data mining}, 2020, pp. 1243--1253.

\bibitem{mcpherson2001birds}
Miller McPherson, Lynn Smith-Lovin, and James~M Cook,
\newblock ``Birds of a feather: Homophily in social networks,''
\newblock {\em Annual review of sociology}, pp. 415--444, 2001.

\bibitem{gerber2013political}
Elisabeth~R Gerber, Adam~Douglas Henry, and Mark Lubell,
\newblock ``Political homophily and collaboration in regional planning
  networks,''
\newblock {\em American Journal of Political Science}, vol. 57, no. 3, pp.
  598--610, 2013.

\bibitem{newman2018networks}
Mark Newman,
\newblock {\em Networks},
\newblock Oxford university press, 2018.

\bibitem{ciotti2016homophily}
Valerio Ciotti, Moreno Bonaventura, Vincenzo Nicosia, Pietro Panzarasa, and
  Vito Latora,
\newblock ``Homophily and missing links in citation networks,''
\newblock {\em EPJ Data Science}, vol. 5, pp. 1--14, 2016.

\bibitem{bruna2013spectral}
Joan Bruna, Wojciech Zaremba, Arthur Szlam, and Yann LeCun,
\newblock ``Spectral networks and locally connected networks on graphs,''
\newblock {\em arXiv preprint arXiv:1312.6203}, 2013.

\bibitem{defferrard2016convolutional}
Micha{\"e}l Defferrard, Xavier Bresson, and Pierre Vandergheynst,
\newblock ``Convolutional neural networks on graphs with fast localized
  spectral filtering,''
\newblock {\em Advances in neural information processing systems}, vol. 29,
  2016.

\bibitem{kipf2016semi}
Thomas~N Kipf and Max Welling,
\newblock ``Semi-supervised classification with graph convolutional networks,''
\newblock {\em arXiv preprint arXiv:1609.02907}, 2016.

\bibitem{hamilton2017inductive}
William~L Hamilton, Rex Ying, and Jure Leskovec,
\newblock ``Inductive representation learning on large graphs,''
\newblock {\em arXiv preprint arXiv:1706.02216}, 2017.

\bibitem{velivckovic2017graph}
Petar Veli{\v{c}}kovi{\'c}, Guillem Cucurull, Arantxa Casanova, Adriana Romero,
  Pietro Lio, and Yoshua Bengio,
\newblock ``Graph attention networks,''
\newblock {\em arXiv preprint arXiv:1710.10903}, 2017.

\bibitem{wu2019simplifying}
Felix Wu, Amauri Souza, Tianyi Zhang, Christopher Fifty, Tao Yu, and Kilian
  Weinberger,
\newblock ``Simplifying graph convolutional networks,''
\newblock in {\em International conference on machine learning}. PMLR, 2019,
  pp. 6861--6871.

\bibitem{qu2019gmnn}
Meng Qu, Yoshua Bengio, and Jian Tang,
\newblock ``Gmnn: Graph markov neural networks,''
\newblock in {\em International conference on machine learning}. PMLR, 2019,
  pp. 5241--5250.

\bibitem{wu2020comprehensive}
Zonghan Wu, Shirui Pan, Fengwen Chen, Guodong Long, Chengqi Zhang, and S~Yu
  Philip,
\newblock ``A comprehensive survey on graph neural networks,''
\newblock {\em IEEE transactions on neural networks and learning systems}, vol.
  32, no. 1, pp. 4--24, 2020.

\bibitem{zhang2020deep}
Ziwei Zhang, Peng Cui, and Wenwu Zhu,
\newblock ``Deep learning on graphs: A survey,''
\newblock {\em IEEE Transactions on Knowledge and Data Engineering}, 2020.

\bibitem{pei2020geom}
Hongbin Pei, Bingzhe Wei, Kevin Chen-Chuan Chang, Yu~Lei, and Bo~Yang,
\newblock ``Geom-gcn: Geometric graph convolutional networks,''
\newblock {\em arXiv preprint arXiv:2002.05287}, 2020.

\bibitem{chamberlain2021grand}
Benjamin~Paul Chamberlain, James Rowbottom, Maria Gorinova, Stefan Webb,
  Emanuele Rossi, and Michael~M Bronstein,
\newblock ``Grand: Graph neural diffusion,''
\newblock {\em arXiv preprint arXiv:2106.10934}, 2021.

\bibitem{chen2020simple}
Ming Chen, Zhewei Wei, Zengfeng Huang, Bolin Ding, and Yaliang Li,
\newblock ``Simple and deep graph convolutional networks,''
\newblock in {\em International Conference on Machine Learning}. PMLR, 2020,
  pp. 1725--1735.

\bibitem{tiwari2022exploring}
Ashish Tiwari, Richeek Das, and Shanmuganathan Raman,
\newblock ``Exploring deeper graph convolutions for semi-supervised node
  classification,''
\newblock in {\em ICASSP 2022-2022 IEEE International Conference on Acoustics,
  Speech and Signal Processing (ICASSP)}. IEEE, 2022, pp. 5463--5467.

\end{thebibliography}

\end{document}